\documentclass[10pt,twocolumn,letterpaper]{article}

\usepackage{cvpr}              

\definecolor{cvprblue}{rgb}{0.21,0.49,0.74}
\usepackage[pagebackref,breaklinks,colorlinks,allcolors=cvprblue]{hyperref}
\newtheorem{thm}{\bf Assumption}
\usepackage{multirow}
\def\paperID{9987} 
\def\confName{CVPR}
\def\confYear{2025}

\usepackage{amsmath}
\usepackage{amssymb}
\usepackage{textcomp}
\usepackage{tablefootnote}
\usepackage{marvosym}
\usepackage[accsupp]{axessibility}

\title{KVQ: Boosting Video Quality Assessment via Saliency-guided Local Perception}

\author{Yunpeng Qu$^{1,2}$, ~Kun Yuan$^{2,\dag(\textrm{\Letter})}$, ~Qizhi Xie$^{1,2}$, ~Ming Sun$^{2}$, Chao Zhou$^{2}$, Jian Wang$^{1,3(\textrm{\Letter})}$ \\
\textsuperscript{\rm 1} Tsinghua University,  \textsuperscript{\rm 2}Kuaishou Technology, \textsuperscript{\rm 3} BNRist, Tsinghua University\\
{\tt \small \{qyp21, xqz20\}@mail.tsinghua.edu.cn,}
{\tt \small jian-wang@tsinghua.edu.cn} \\
{\tt \small \{yuankun03,sunming03,zhouchao\}@kuaishou.com}}

\begin{document}
\maketitle

\renewcommand{\thefootnote}{}
\footnotetext{$^{\dag}$ Project leader. $^{\textrm{\Letter}}$ Corresponding authors.}

\begin{abstract}
Video Quality Assessment (VQA), which intends to predict the perceptual quality of videos, has attracted increasing attention.
Due to factors like motion blur or specific distortions, the quality of different regions in a video varies.
Recognizing the region-wise local quality within a video is beneficial for assessing global quality and can guide us in adopting fine-grained enhancement or transcoding strategies.
Due to the heavy cost of annotating region-wise quality, the lack of ground truth constraints from relevant datasets further complicates the utilization of local perception.
Inspired by the Human Visual System (HVS) that links global quality to the local texture of different regions and their visual saliency, we propose a Kaleidoscope Video Quality Assessment (KVQ) framework, which aims to effectively assess both saliency and local texture, thereby facilitating the assessment of global quality.
Our framework extracts visual saliency and allocates attention using Fusion-Window Attention (FWA) while incorporating a Local Perception Constraint (LPC) to mitigate the reliance of regional texture perception on neighboring areas.
KVQ obtains significant improvements across multiple scenarios on five VQA benchmarks compared to SOTA methods.
Furthermore, to assess local perception, we establish a new Local Perception Visual Quality (LPVQ) dataset with region-wise annotations.
Experimental results demonstrate the capability of KVQ in perceiving local distortions.
KVQ models and the LPVQ dataset will be available at \url{https://github.com/qyp2000/KVQ}.
\end{abstract}

\section{Introduction}
\label{sec:intro}

In the era of burgeoning video content-driven social media platforms, there has been an unprecedented surge in the creation and dissemination of videos.
To enhance users' Quality of Experience (QoE), Video Quality Assessment (VQA), which aims to forecast the human perceptual quality of a video, 
has attracted raising attention \cite{DBLP:journals/tip/TuWBAB21}.
VQA facilitates the identification of low-quality videos, thereby guiding the video enhancement and encoding system \cite{DBLP:conf/cvpr/ZhangIESW18, DBLP:conf/cvpr/ChadhaA21}, resulting in a superior visual experience while effectively reducing bandwidth costs.
Due to the difficulty in obtaining reference videos in most user-generated content scenarios, we focus on the No-Reference (NR) VQA domain \cite{DBLP:conf/iccv/KeWWMY21, DBLP:journals/tcsv/WuCLHSYL23, DBLP:journals/ijcv/LiJJ21}.

\begin{figure}[t]
  \centering
  \includegraphics[width=0.9\linewidth]{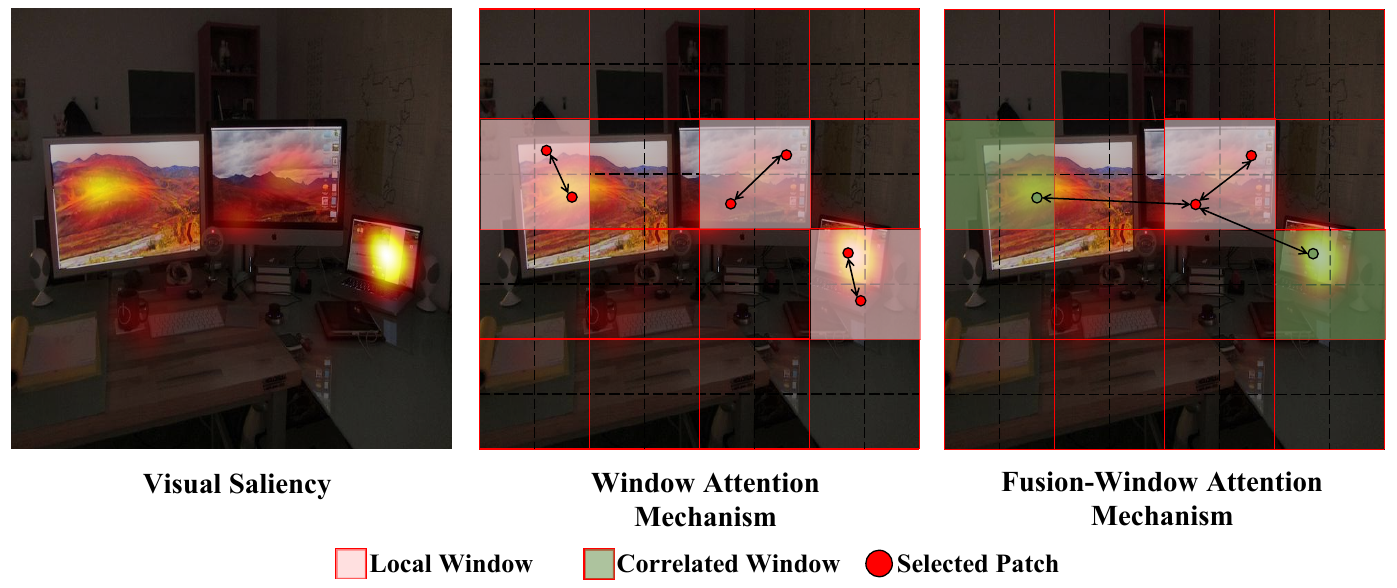}
  \caption{Human visual saliency and different attention mechanisms. (\textit{Left}) Visual saliency obtained through mouse-tracking in the SALICON \cite{jiang2015salicon} dataset. (\textit{Middle}) Window Attention mechanism \cite{DBLP:conf/iccv/LiuL00W0LG21}, where patches establish connections only within the window. (\textit{Right}) Our Fusion-Window Attention, where patches can allocate long-range attention across correlated regions. }
  \label{fig:fwa}
\end{figure}

Classical VQA algorithms rely on handcrafted features to predict quality \cite{DBLP:journals/tip/MittalMB12, DBLP:journals/tip/Korhonen19, DBLP:journals/tip/TuWBAB21}. In recent years, numerous deep learning methods based on CNN \cite{DBLP:journals/ijcv/LiJJ21, DBLP:conf/mm/LiJJ19,zhao2023quality} or Transformer \cite{,DBLP:journals/tcsv/WuCLHSYL23, xie2024qpt} have also been applied 
 \cite{DBLP:journals/tcsv/LiZTZW22, DBLP:conf/cvpr/YingMGB21}.
These methods focus on predicting the global quality of videos.
Indeed, there are differences in the quality of disparate spatiotemporal regions within a video due to their distinct textures and distortions (\textit{e.g.}, motion blur, compression), and it may not align with the global perceptual quality \cite{DBLP:journals/tip/MaLZDWZ18}.
Achieving a perception of region-wise distortions is crucial for VQA.
Nevertheless, utilizing local perception to assist VQA poses a challenge, owing to the dearth of publicly accessible datasets meeting the constraints of local perception.
It is a time-consuming and complicated task to annotate Mean Opinion Scores (MOS) for videos, as it requires a large number of participants to ensure reliability \cite{lu2024kvq, liu2023ada, yuan2024ptm}.
Annotating the local quality of spatiotemporal regions in videos incurs even more challenging costs, thereby escalating the annotation expenses by approximately $\mathcal{O}(N^3)$.
This renders the acquisition of extensive local quality annotations for training nearly impracticable.
Several approaches \cite{DBLP:conf/eccv/WuCHLWSYL22, DBLP:conf/cvpr/YingMGB21} attempt to predict local quality by outputting maps.
However, these methods lack robust constraints on local perception, resulting in quality maps influenced by visual saliency.
Current methods can be further refined to enhance local quality perception.

Research on the Human Visual System (HVS) has revealed that global quality is influenced by both the visual saliency and local texture of individual regions. 
\cite{DBLP:journals/tnn/ZhangBWCL16, DBLP:journals/tip/ZhangL17}.
\textit{\textbf{Visual saliency} captures the allocation of human visual attention, with regions exhibiting key semantic features, high contrast, or significant differences from the surrounding areas being more salient} \cite{bruce2005saliency, treue2003visual}.
Specifically, visual saliency encompasses the understanding of scene semantics and the correlation between regions.
\textit{\textbf{Local texture} refers to the low-level visual features within the region, such as details, texture patterns, color variations, and distortions, without considering higher-level semantic content }\cite{haralick1973textural}.
To obtain a better assessment, it is necessary to make reliable predictions for both saliency and local texture.
As shown in Fig.~\ref{fig:fwa}, the mouse-tracking ground truth reveals that visual saliency involves the global allocation of attention.
Therefore, we believe that enabling models to adaptively allocate global-wise attention across correlated regions, akin to human vision, can facilitate the extraction of visual saliency.
Furthermore, facing the limited availability of local annotations, we consider leveraging priors to bolster the reliability of local perception, \textit{i.e.}, the local texture of a region is determined solely by the distortions within its specific areas and is not influenced by other areas.



Based on the principles above, we propose a \textit{Kaleidoscope Video Quality Assessment (KVQ)} framework in this paper. Structurally, a video Transformer serves as the backbone, upon which a dual branch is employed to forecast both the saliency map and the localized texture map.
To achieve precise saliency extraction, we present a \textit{Fusion-Window Attention (FWA)}, which selectively allocates global-wise visual attention.
To bolster the reliability of local perception, we devise a \textit{Local Perception Constraint (LPC)}, which aims to mitigate the dependency of regional texture perception on neighboring regions. 
Our contributions are as follows:
\begin{itemize}
    \item 
    Inspired by the HVS, we make two assumptions, enhancing local perception and saliency extraction for VQA.
    \item We propose the KVQ framework, which incorporates the FWA module for adaptive global-wise saliency extraction and the LPC for perceiving local texture. 
    Our model is capable of independently predicting visual saliency and local perception for each region in a video.
    \item KVQ achieves SOTA results compared to existing methods in three evaluation scenarios: intra-dataset, cross-dataset, and transfer learning evaluation. Extensive ablation studies prove the validity of each component.
    \item To verify the assessment of local distortions and texture, we establish a new \textit{Local Perception Visual Quality (LPVQ)} dataset with region-wise annotations. 
    It comprises 50 images annotated by 14 visual experts, totaling 34,300 annotations. 
    Experimental results show the strong capability of KVQ in local perception.
\end{itemize}

\section{Related Works}
\label{sec:related}

\paragraph{\textbf{\textit{Visual Saliency in VQA.}}}
In the fields of VQA or IQA, many methods incorporate visual saliency measures to capture attention variations for better quality prediction.
 \cite{DBLP:conf/mm/YuanKZSW23, DBLP:journals/tcsv/WuCLHSYL23, DBLP:journals/tmm/GuanYZCW17, DBLP:journals/tip/BosseMMWS18}.
These methods often lack clear definitions or constraints on the saliency.
SGDNet \cite{DBLP:conf/mm/YangJLW19}, TranSLA \cite{DBLP:conf/iccvw/ZhuHCXLC21}, and MMMNet \cite{DBLP:journals/tcsv/LiZC21} incorporate saliency prediction as a subtask of quality assessment.
However, these methods have complex training processes and rely on the predictions of SOTA saliency models as ground truth.
Due to the attention mechanism's capacity to reflect region correlations, exploring effective attention mechanisms to extract visual saliency is valuable.
Many previous works have excelled in various visual tasks by employing handcrafted attention patterns, such as window attention \cite{DBLP:conf/iccv/LiuL00W0LG21}, dilated attention \cite{DBLP:conf/iclr/0001YCL00L22}, and deformable attention \cite{DBLP:conf/cvpr/XiaPSLH22}.
However, these attention computations are still confined to pre-designed static patterns in neighboring areas, limiting long-term attention allocation.
Therefore, we propose a cross-regional attention mechanism to adaptively capture global visual saliency.

\paragraph{\textbf{\textit{Local perception in VQA.}}}
HVS tends to perceive the inherent characteristics of a video from both a spatial and temporal perspective \cite{DBLP:conf/mm/SunMLZ22, DBLP:journals/corr/topiq, DBLP:journals/corr/rapique}. 
The subjective quality of regions within a frame exhibit significant variations \cite{DBLP:conf/cvpr/HosuGS19, DBLP:conf/iccv/KeWWMY21} and the global quality can be determined by a few distorted key frames \cite{DBLP:conf/mm/YuanKZSW23, DBLP:conf/cvpr/ZhaoYSW23}. 
Previous works have harnessed this prior \cite{DBLP:conf/eccv/WuCHLWSYL22, DBLP:conf/cvpr/YingMGB21}. 
Fast-VQA \cite{DBLP:conf/eccv/WuCHLWSYL22, DBLP:journals/corr/abs-2210-05357} uniformly samples local spatial patches throughout the video and computes their respective quality. PVQ \cite{DBLP:conf/cvpr/YingMGB21} divides a video into multiple forms of local patches and annotates them through crowd-sourced studies.
However, the above methods lack region-wise human annotations and intertwine local texture with local visual saliency, which poses challenges for them to effectively perceive the quality of different regions in a video.
In this study, we explicitly decouple the saliency and texture maps without relying on any localized supervision signal, thereby enhancing performance and interpretability.

\section{Methods}
\label{sec:methods}

\subsection{Revisiting Video Transformer}
\label{method:swin}
The video Transformer is the core framework in our paper, capturing spatiotemporal information.
In this architecture, attention blocks \cite{DBLP:conf/nips/VaswaniSPUJGKP17} are relied upon to model interdependencies across positions
Given queries $\mathbf{Q} \in \mathbb R^{N_q\times C}$, keys $\mathbf{K} \in \mathbb R^{N_k\times C}$, and values $\mathbf{V} \in \mathbb R^{N_v\times C}$, an attention function can be denoted as follows and are typically applied in the form of Multi-head Self-Attention (MSA): 
\begin{equation}
\label{eq:attention}
    Attn(\mathbf{Q}, \mathbf{K}, \mathbf{V}) = Softmax(\frac{\mathbf{Q}\mathbf{K}^T}{\sqrt{C}})\mathbf{V}.\\
\end{equation}

For a video $\mathcal{V} \in \mathbb R^{T_v\times H_v\times W_v\times 3}$ with $T_v$ frames of size $H_v\times W_v$, a video Transformer reshapes the video clip into non-overlapping patches $\mathbf{X}\in \mathbb R^{T\times H\times W\times 3P_tP^2}$ for processing, where $(P_t, P, P)$ represents the patch size and $(T, H, W)$ denotes the number of patches across spatiotemporal dimensions.
The backbone network includes stacked blocks with MSA and Feed-Forward Network (FFN) layers. In the $l$-th block, the computation is as follows:
\begin{equation}
\label{eq:block}
    \mathbf{X}^{l^{\prime}} = \mathcal{MSA}(\mathbf{X}^l),
    \mathbf{X}^{l+1} =\mathcal{FFN}(\mathbf{X}^{l^{\prime}}).
\end{equation}
In our paper, the Video Swin Transformer (Swin-T) \cite{DBLP:conf/cvpr/LiuN0W00022} is selected for its capability to retain spatiotemporal features.

\subsection{HVS-based Visual Perception}
\label{method:theorm}

\begin{figure}[t]
  \centering
  \includegraphics[width=0.95\linewidth]{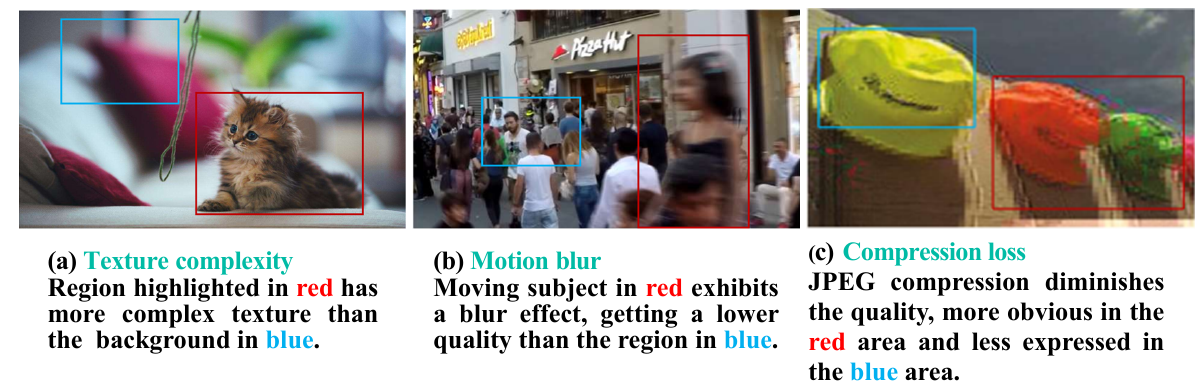}
  \caption{Perceptual quality of different regions varies due to factors (\textit{e.g.}, texture, motion, compression). These variations may not necessarily align with the global quality.}
  \label{fig:loacalquality}
\end{figure}
In Fig.~\ref{fig:loacalquality}, the perceptual quality of different spatiotemporal regions in a video may vary due to multiple factors, such as:
\begin{itemize}
    \item \textbf{Texture complexity}. 
    Regions may exhibit simple, uniform textures or intricate details, causing quality variations during encoding or compression.
    \item \textbf{Motion complexity}. 
    Rapid motion or sudden changes in the scene can result in motion blur, artifacts, or detail loss, leading to varying perceptual quality.
    \item \textbf{Compression parameters}. 
    Bitrate allocation varies across frames or regions, prioritizing complex areas for quality preservation and leading to quality discrepancies due to non-uniform allocation.
\end{itemize}

For a precise assessment of the correlation between local quality and global quality, we can draw inspiration from the HVS to address this issue: \textit{\textbf{What factors in a video influence human perceptual quality?}}

\textbf{Firstly}, human attention allocation to different regions is inherently uneven \cite{DBLP:journals/tmm/GuWYLZY016}, with video perceptual quality predominantly determined by intensely focused regions or frames \cite{DBLP:conf/mm/YangJLW19, DBLP:conf/cvpr/ZhaoYSW23}.
These visually salient and focused regions may encompass noteworthy semantic information, exhibit high contrast that stands out from other regions, or exhibit significant differences from neighboring areas \cite{bruce2005saliency, treue2003visual}.
This \textbf{visual saliency} reflects how the perceived quality of regions is influenced by high-level semantics and other areas.

\textbf{Secondly}, besides high-level semantics and inter-regional correlations, the low-level features inherent to a region also influence perceptual quality.
We define these features as \textbf{local texture}, reflecting the inherent distortions and low-level attributes, such as brightness, sharpness, and intricate patterns \cite{wang2010information, haralick1973textural}.
As two distinct factors influencing quality, the local texture should be entirely decoupled from visual saliency, free from any semantic or regional correlations, as these high-level understandings are manifested in saliency.
Hence, local texture reflects the intrinsic characteristics within a region, uninfluenced by other regions.

Numerous studies on the HVS have indicated that the global quality is influenced by both local texture features and visual saliency \cite{wang2010information}.
Building upon these mechanisms, we denote the regions in $\mathcal{V}$ as $\{\mathbf{x}_{i,j,k}| 0\le i\le T, 0\le j\le H,  0\le j\le W \}$ and make the first assumption:

\begin{thm}\label{thm:saliency}
    The comprehensive quality of a video is acquired by assigning weightage to the local texture based on the distribution of visual saliency.
    Defining the computed saliency map as $\mathcal{S} \in \mathbb{R}^{T\times H\times W}$ and the local texture map as $\mathcal{Q} \in \mathbb{R}^{T\times H\times W}$, the predicted quality $q$ can be expressed as the weighted average of the element-wise multiplication:
\begin{equation}
\label{eq:quality}
    q = \frac{1}{THW}\sum_{i}^T\sum_{j}^H\sum_{k}^W\mathcal{S}_{i,j,k}\cdot \mathcal{Q}_{i,j,k}.
\end{equation}
\end{thm}

\begin{figure*}[t]
  \centering
    \includegraphics[width=0.92\linewidth]{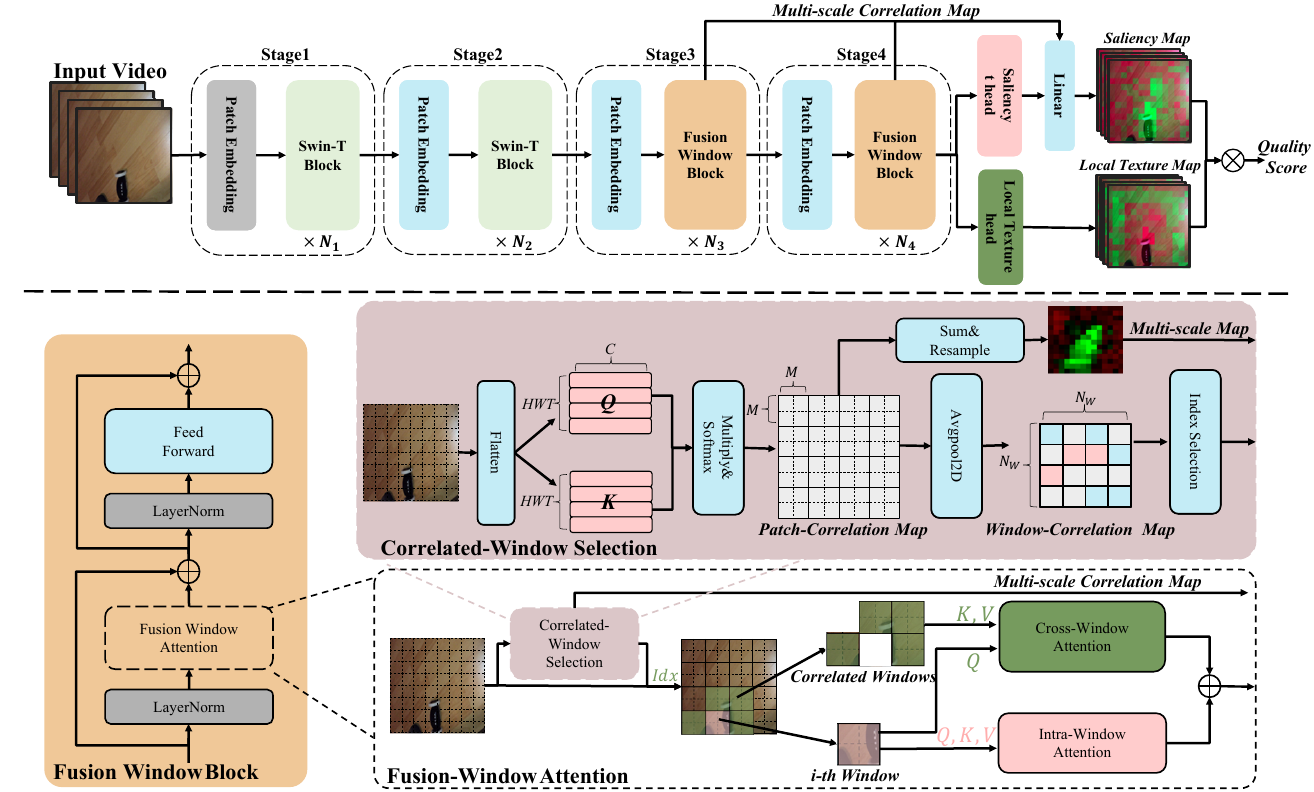}
  \caption{The framework of our proposed KVQ.}
  \label{fig:model}
\end{figure*}

However, relying solely on \textit{Assumption \ref{thm:saliency}} is not sufficient, as it can not effectively distinguish between visual saliency and local texture.
Local texture reflects the local low-quality factors and distortions, which are what we aim to predict in the local perception.
Given the limited availability of local annotations in VQA datasets, it is crucial to establish suitable constraints utilizing prior information to bolster the reliability of local perception. 
Based on the previous analysis, local texture captures only the detailed characteristics specific to that region, such as distortions and sharpness, without considering the influence of other areas.
Building upon the criteria for annotating local quality \cite{DBLP:journals/jstsp/Winkler12,DBLP:conf/qomex/HosuHJLMSLS17}, we can make the second assumption:
\begin{thm}\label{thm:local}
The local texture of a region is determined solely by the distortions present within a specific area and is not influenced by other areas. Hence, evaluating distortions of an individual region by feeding it separately to the VQA model $\mathcal{F}(\cdot)$ should yield consistent results with the corresponding region by feeding the entire video as input.
\begin{equation}
\label{eq:local}
    \mathcal{Q}_{i,j,k} \sim \mathcal{F}(\mathbf{x}_{i,j,k}), ~\text{where} ~ \mathcal{Q}_{i,j,k} = \mathcal{F}(\mathbf{X})_{i,j,k}.
\end{equation}
\end{thm}
Classical works like SSIM \cite{wang2010information, wang2004image} adopt similar definitions, partitioning images into patches and considering low-level features limited to each patch as local quality, which aligns with our \textit{Assumption \ref{thm:local}} by considering local textures independently of correlations. 
In this paper, we decouple the prediction of saliency and local texture based on the above assumptions and achieve effective modeling of local perception by establishing additional constraints.

\subsection{KVQ Framework}
\label{method:overview}
In Fig. \ref{fig:model}, we propose the KVQ framework, leveraging a Video Swin-T backbone for generating video features.
\textit{According to Assumption \ref{thm:saliency}}, at the forefront, a dual branch renders predictions respectively for the saliency map and local texture map, combined through weighting to derive the quality score.
To enhance global-wise saliency extraction, we propose a Fusion-Window Attention (FWA) module. 
To bolster local perception, \textit{in accordance with Assumption \ref{thm:local}}, we formulate a Local Perception Constraint (LPC) to eliminate the reliance of regional perception on adjacent regions.


\subsubsection{\textbf{Fusion-Window Attention}} 
Visual saliency entails allocating attention globally, advocating for long-range connections between regions instead of restricting attention to local areas.
Inspired by the mechanism of \cite{DBLP:conf/cvpr/ZhuWKZL23}, which encourages each query to focus on semantically relevant key-value pairs, we believe such cross-regional attention modeling aids in capturing global saliency for the VQA task.
Thus, we extend this idea to the VQA task and further propose a correlation-aware Fusion-Window Attention module.
We incorporate an adaptive selection of correlated windows for cross-window attention in addition to the attention within each window. 
This fusion of intra-window and cross-window attention enables patches to allocate attention to all relevant patches in a global scope.
As shown in Fig.~\ref{fig:model}, the procedure is as follows:


\paragraph{\textbf{\textit{Correlated-Window Selection.}}}
There is an initial module for Correlated-Window Selection (CWS), which adaptively selects the most correlated windows.
Given an input feature map $\mathbf{X}^l \in \mathbb R^{T_l\times H_l\times W_l\times C}$ in the $l$-th block, we first divide it into non-overlapping windows $\mathbf{X}^l_w \in \mathbb R^{N_w \times M\times C}$, with $M$ denoting the patches per window and $N_w$ indicating the total window count. 
$\mathbf{X}^l_w$ are then mapped into queries, keys and values as $\mathbf{Q},\mathbf{K},\mathbf{V} \in \mathbb R^{N_w \times M\times C}$ through projections: 
\begin{equation}
    \begin{split}
    \mathbf{Q} = \mathbf{X}^l_w\mathbf{W}^q_c,\ \mathbf{K} = \mathbf{X}^l_w\mathbf{W}^k_c, \ \mathbf{V} = \mathbf{X}^l_w\mathbf{W}^v_c,
    \end{split}
\end{equation}
where $\mathbf{W}^q_c, \mathbf{W}^k_c, \mathbf{W}^v_c \in \mathbb R^{C \times C}$ are projection weights.

Subsequently, we flatten the patches and establish the patch-correlation map $\mathbf{I}_p \in \mathbb R^{N_wM\times N_wM}$ using the attention mechanism.
The patch-level map is applied pooling to calculate the correlations between windows, resulting in a window-level correlation map $\mathbf{I}_w \in \mathbb R^{N_w\times N_w}$.
We select the top-$k$ windows that have the highest correlation, obtaining the indices of these windows $\mathbf{Idx}\in \mathbb R^{N_w\times k}$:
\begin{equation}
    \begin{split}
    \mathbf{I}_p = Soft&max(Flatten(\mathbf{Q}) \cdot Flatten(\mathbf{K})^T), \\
    &\mathbf{I}_w = Avgpool2D(\mathbf{I}_p),\\
    &\mathbf{Idx} = TopkMax(\mathbf{I}_w,k),
    \end{split}
\end{equation}
where each row in $\mathbf{Idx}$ has $k$ indices of correlated windows.

\paragraph{\textbf{\textit{Intra-window and Cross-window Attention.}}}
Given the semantic similarity among neighboring regions and their influence on human visual attention, we maintain self-attention within each window, known as Intra-Window Attention (IWA).
In addition, based on the correlation indices obtained from the CWS module, we utilize Cross-Window Attention (CWA) to establish inter-window connectivity.

In CWA, we gather the keys and values of the correlated windows based on $\mathbf{Idx}$.
Each window provides queries, while the correlated windows provide keys and values.
The computation of the CWA is as follows:
\begin{equation}
    \begin{split}
    \mathcal{CWA}(\mathbf{X}^l) = Attn(\mathbf{Q}, \mathbf{K}[\mathbf{Idx}], \mathbf{V}[\mathbf{Idx}]),
    \end{split}
\end{equation}

Similarly, we can also denote the computation process of the IWA. 
The final result of the FWA module is the sum:
\begin{equation}
    \begin{split}
    \mathcal{IWA}(\mathbf{X}^l)& = Attn(\mathbf{X}^l_w\mathbf{W}^q_i,\ \mathbf{X}^l_w\mathbf{W}^k_i, \ \mathbf{X}^l_w\mathbf{W}^v_i),\\
    \mathcal{FWA}&(\mathbf{X}^l) = \mathcal{IWA}(\mathbf{X}^l) + \mathcal{CWA}(\mathbf{X}^l),
    \end{split}
\end{equation}
where $\mathbf{W}^q_i, \mathbf{W}^k_i, \mathbf{W}^v_i \in \mathbb R^{C \times C}$ are projection weights.
Compared to images, videos inherently possess temporal correlations between consecutive frames.
To effectively model these spatiotemporal characteristics, our KVQ framework adopts the video-specific Swin-T architecture. 
Particularly through the FWA module, KVQ captures long-range temporal dependencies spanning multiple frames by establishing global attention connections across windows. 
Therefore, our KVQ can effectively characterize the video-specific temporal correlation patterns.

\paragraph{\textbf{\textit{Multi-scale Ensemble Saliency Map.}}}
Inspired by the HVS where visual information flows through a cortical hierarchy \cite{wurtz2000central}, we ensemble the multi-scale correlation maps acquired from $N$ block for the final saliency map, as $\mathbf{I}_p$ also reflect the allocation of visual attention. 
By transposing $\mathbf{I}_p$ and summing each column, we obtain $\mathbf{I}_p^{'} \in \mathbb R^{N_wM}$, where each element reflects the significance of each patch.
In the $l$-th block, $\mathbf{I}_p'$ is reshaped and pooled to form $\mathbf{I}_p^{(l)} \in \mathbb R^{T\times H \times W}$, which is then fused with the saliency output $\widetilde{\mathcal{S}}$:
\begin{equation}
    \begin{split}
    \mathbf{I}_p' &= Sum(\mathbf{I}_p^T),\\
    \mathbf{I}_p^{(l)} = Avg&pool3D(Resample(\mathbf{I}_p')), \\
    \mathcal{S} = Softma&x\big(w^0\widetilde{\mathcal{S}}+\sum\nolimits_{l=1}^N w^l \cdot \mathbf{I}_p^{(l)}\big),
    \end{split}
\end{equation}
where $w^l$ is the weight for balancing.

\subsubsection{\textbf{Local Perception Constraint}}
\begin{figure}[t]
  \centering
    \includegraphics[width=0.9\linewidth]{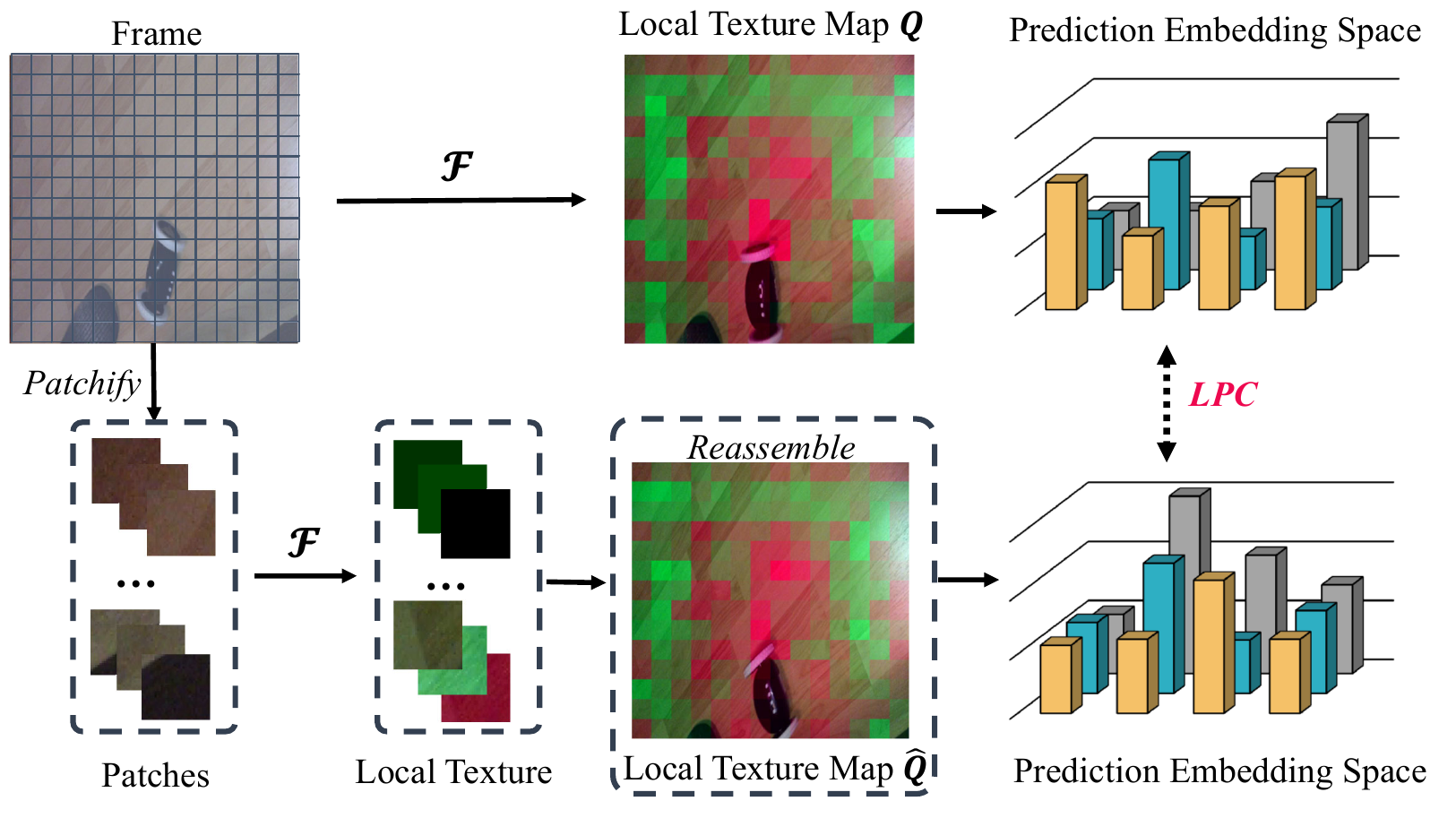}
  \caption{Illustration of Local Perception Constraint.}
  \label{fig:patchloss}
\end{figure}

In addition to the saliency map $\mathcal{S}$, a local texture map $\mathcal{Q}$ is also generated. 
Based on the \textit{Assumption \ref{thm:local}}, local texture reflects low-level features of different regions, independent of the global semantics and inter-regional correlations.
We propose that the LPC facilitates the features to focus exclusively on the distortions inherent within the particular region, thereby minimizing the influence of visual saliency and enhancing local perception.
As shown in Fig.~\ref{fig:patchloss}, video $\mathbf{X}$ is fed into $\mathcal{F}$ to obtain texture map $\mathcal{Q}$, while $\mathbf{X}$ is sliced into patches $\mathbf{x}_{i,j,k}$, each fed into $\mathcal{F}$ separately, and reassembled to get $\hat{\mathcal{Q}}$.
According to Eq.~\ref{eq:local}, this constraint aims to minimize the distance between prediction embeddings obtained in two different ways: 
\begin{equation}\small
    \begin{split}
    & \mathcal{Q}_{i,j,k} = \mathcal{F}(\mathbf{X})_{i,j,k}, ~ \hat{\mathcal{Q}}_{i,j,k} = \mathcal{F}(\mathbf{x}_{i,j,k }), \\
    & \mathcal{L}_{lpc} = 1-\frac{ \sum_{i}^T\sum_{j}^H\sum_{k}^W  \mathcal{Q}_{i,j,k}\cdot \hat{\mathcal{Q}}_{i,j,k}}{||\mathcal{Q}||\cdot ||\hat{\mathcal{Q}}||}.
    \end{split}
\end{equation}


\subsection{Optimization Objective}
\label{method:loss}
We utilize the commonly used PLCC loss as the primary loss function $\mathcal{L}_{plcc}$ \cite{DBLP:conf/mm/LiJJ20}. 
Learning the relative quality relationship between videos is a crucial means to enhance robustness \cite{DBLP:journals/corr/abs-2210-05357}. 
We incorporate the rank loss, where the predictions $q$ and the ground truth $q_{gt}$ are compared in terms of their ranking, as another loss $\mathcal{L}_{rank}$.
The final optimization objective is a weighted combination of the two losses mentioned above and the LPC:
\begin{equation}
     min~ \mathcal{L}_{plcc} + \lambda_{r}\mathcal{L}_{rank} + \lambda_{p}\mathcal{L}_{lpc}. \\
\end{equation}
where $\lambda_{r}$ and $\lambda_{p}$ are balancing coefficients.


\section{Experiments}
\label{sec:experiments}
\subsection{Experimental Setups}

\paragraph{\textbf{\textit{Dataset and Evaluation Metrics.}}} 
We utilize the large-scale LSVQ\textsubscript{train} \cite{DBLP:conf/cvpr/YingMGB21} dataset (with 28,056 videos) for training and validation on the intra-dataset test subsets: LSVQ\textsubscript{test} and LSVQ\textsubscript{1080p}. 
Besides, three widely-recognized smaller benchmarks are used for cross-dataset and transfer learning evaluation, including KoNViD-1k \cite{DBLP:conf/qomex/HosuHJLMSLS17}, LIVE-VQC \cite{DBLP:journals/tip/SinnoB19}, and YouTube-UGC \cite{DBLP:conf/mmsp/WangIA19}. 
LIVE-VQC contains 585 videos ranging in resolution from 240P to 1080P. 
KoNViD-1k comprises 1,200 videos with a resolution of $960\times540$, sampled from YFCC100M \cite{DBLP:journals/cacm/ThomeeSFENPBL16}.
We use Mean Opinion Scores (MOS) to represent the subjective quality scores and employ PLCC and SRCC as metrics for evaluation.

\begin{figure}[t]
  \centering
    \includegraphics[width=0.87\linewidth]{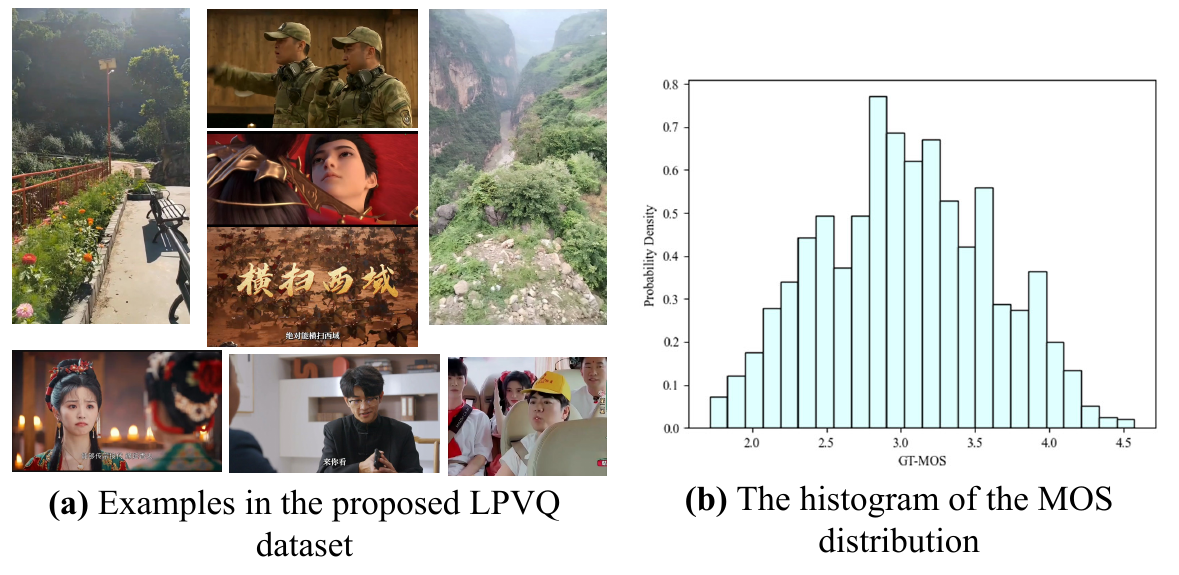}
  \caption{Examples and the MOS distribution in the proposed LPVQ dataset. Please zoom in for a better view.}
  \label{fig:lpvq}
\end{figure}

\begin{table*}[t]
  \centering
  \footnotesize
  \caption{Comparison with SOTA methods under the intra-dataset and generalization settings. The best scores are colored in \textcolor{red}{red}.}
  \label{tab:sota}
  \begin{tabular}{c|c|cc|cc|cc|cc}
    \toprule
    Training Set: LSVQ\textsubscript{train} & Computational Cost & \multicolumn{4}{c|}{Intra-dataset Evaluations} & \multicolumn{4}{c}{Generalization Evaluations} \\
    \midrule
    \multirow{2}{*}{Method} & \multirow{2}{*}{GFlops} &\multicolumn{2}{c|}{LSVQ\textsubscript{test}} & \multicolumn{2}{c|}{LSVQ\textsubscript{1080p}} & \multicolumn{2}{c|}{KoNViD-1k} & \multicolumn{2}{c}{LIVE-VQC}  \\
    & & SRCC & PLCC & SRCC & PLCC & SRCC & PLCC & SRCC & PLCC  \\ 
    \midrule
    BRISQUE \cite{DBLP:journals/tip/MittalMB12} & NA & 0.569 & 0.576 &0.497 &0.531 &0.646 &0.647 &0.524 &0.536 \\
    TLVQM \cite{DBLP:journals/tip/Korhonen19} & NA &0.772 &0.774 &0.589 &0.616 &0.732 &0.724 &0.670 &0.691  \\
    VIDEVAL \cite{DBLP:journals/tip/TuWBAB21} & NA &0.794 &0.783 &0.545& 0.554 &0.751 &0.741 &0.630 &0.640 \\
    VSFA \cite{DBLP:conf/mm/LiJJ19} &  40919 &0.801 &0.796 &0.675 &0.704 &0.784 &0.794 &0.734 & 0.772\\
    PVQ\textsubscript{wo/patch} \cite{DBLP:conf/cvpr/YingMGB21} &58501 &0.814 &0.816 &0.686 &0.708 &0.781 &0.781 &0.747 &0.776  \\
    PVQ\textsubscript{w/patch} \cite{DBLP:conf/cvpr/YingMGB21} & 58501 &0.827 &0.828&0.711 &0.739 &0.791 &0.795 &0.770 &0.807  \\
    Li \textit{et al.} \cite{DBLP:journals/tcsv/LiZTZW22} & 112537 & 0.852 & 0.855 &0.771 &0.782& 0.834& 0.837 &0.816 &0.824\\
    CLIPVQA \cite{xing2024clipvqa} & 658 &0.883 & 0.885 & 0.785 & 0.833& NA&NA & NA&NA\\
    
    Fast-VQA \cite{DBLP:conf/eccv/WuCHLWSYL22} & 279.1 & 0.876 &0.877 &0.779 &0.814 &0.859 &0.855 &\textcolor{red}{\textbf{0.823}} &\textcolor{red}{\textbf{0.844}}  \\
    Faster-VQA \cite{DBLP:journals/corr/abs-2210-05357} & 69.8 & 0.873 & 0.874 & 0.772 & 0.811 & 0.863 & 0.863 & 0.813 & 0.837 \\
    \midrule
    \textbf{KVQ} & 353.0  &  \textcolor{red}{\textbf{0.896}} & \textcolor{red}{\textbf{0.897}} & \textcolor{red}{\textbf{0.814}} & \textcolor{red}{\textbf{0.846}} & \textcolor{red}{\textbf{0.890}} & \textcolor{red}{\textbf{0.892}} & 0.820 & 0.843\\
    \midrule
    \textit{Improvement to Fast-VQA} & \textit{+26.5\%} & \textit{+2.3\%}&\textit{+2.3\%} &\textit{+4.5\%}&\textit{+3.9\%} &\textit{+3.6\%}&\textit{+4.3\%}& \textit{-0.4\%} & \textit{-0.1\%}
    \\
    \bottomrule
  \end{tabular}
\end{table*}
\begin{table*}[t]
  \centering
  \footnotesize
  \caption{Comparison with SOTA methods using transfer learning. The best scores are colored in \textcolor{red}{red}.}
  \begin{tabular}{c|c|cc|cc|cc}
    \toprule
    \multicolumn{2}{c|}{Target Dataset}  & \multicolumn{2}{c|}{LIVE-VQC} & \multicolumn{2}{c|}{KoNViD-1k} & \multicolumn{2}{c}{YouTube-UGC} \\
    \midrule
    Method & Pre-training Dataset & SRCC & PLCC & SRCC & PLCC & SRCC & PLCC  \\ 
    \midrule
    
TLVQM \cite{DBLP:journals/tip/Korhonen19}& NA (pure handcraft) &0.799 &0.803 &0.773 &0.768 &0.669 &0.659 \\
VIDEVAL  \cite{DBLP:journals/tip/TuWBAB21} &NA (pure handcraft)  &0.752 & 0.751 & 0.783 & 0.780 & 0.779  &0.773\\
\midrule
VSFA \cite{DBLP:conf/mm/LiJJ19} & None  &0.773  &0.795 & 0.773  &0.775  &0.724  &0.743 \\
PVQ  \cite{DBLP:conf/cvpr/YingMGB21} & PaQ-2-PiQ \cite{DBLP:conf/cvpr/YingNGMGB20}  &0.827  &0.837  &0.791  &0.786  &NA  &NA \\
CoINVQ  \cite{DBLP:conf/cvpr/WangKTYBAMY21} & self-collected  &NA  &NA  &0.767  &0.764  &0.816  &0.802\\

 \multirow{2}{*}{Li \textit{et al.}\cite{DBLP:journals/tcsv/LiZTZW22}} & BID \cite{DBLP:journals/tip/CiancioCSSSO11}, LIVE \cite{DBLP:journals/tip/GhadiyaramB16}, & \multirow{2}{*}{0.834} &\multirow{2}{*}{0.842} &\multirow{2}{*}{0.834}& \multirow{2}{*}{0.836} &\multirow{2}{*}{0.818} &\multirow{2}{*}{0.826} \\
~ & KonIQ-10k \cite{DBLP:journals/tip/HosuLSS20}, SPAQ \cite{DBLP:conf/cvpr/FangZZMW20} &&&& & & \\

DisCoVQA \cite{DBLP:journals/tcsv/WuCLHSYL23}& None &0.820 &0.826&0.847& 0.847 & NA&NA\\
2BiVQA \cite{DBLP:conf/mm/YuanKZSW23} & ImageNet \cite{DBLP:conf/cvpr/DengDSLL009}, KonIQ-10k \cite{DBLP:journals/tip/HosuLSS20} & 0.761 &0.832 &0.815& 0.835 & 0.771 &0.790\\
VQT\cite{DBLP:conf/mm/YuanKZSW23} & ImageNet \cite{DBLP:conf/cvpr/DengDSLL009}, Kinetics-400 \cite{DBLP:journals/corr/KayCSZHVVGBNSZ17} &0.824 &  0.836  & 0.858 & 0.8684 & 0.836 & 0.851\\
StarVQA \cite{xing2022starvqa} & LSVQ \cite{DBLP:conf/cvpr/YingMGB21}   & 0.753& 0.809 &0.842 &0.849& NA&NA \\
StarVQA+ \cite{xing2023starvqa+} &   ImageNet \cite{DBLP:conf/cvpr/DengDSLL009}, LSVQ \cite{DBLP:conf/cvpr/YingMGB21}, ... & 0.857 & 0.874& 0.876&0.881 &0.826&0.820 \\

Fast-VQA  \cite{DBLP:conf/eccv/WuCHLWSYL22} & LSVQ \cite{DBLP:conf/cvpr/YingMGB21}   &0.849  &0.862  &0.891  &0.892  &0.855  &0.852 \\
FasterVQA \cite{DBLP:journals/corr/abs-2210-05357} & LSVQ \cite{DBLP:conf/cvpr/YingMGB21}   & 0.843 & 0.858 &0.895 &0.898  &0.863 &0.859 \\ 
    \midrule
    \textbf{KVQ} & LSVQ \cite{DBLP:conf/cvpr/YingMGB21}   &\textcolor{red}{\textbf{0.859}} &\textcolor{red}{\textbf{0.879}}  & \textcolor{red}{\textbf{0.909}} &\textcolor{red}{\textbf{0.915}}  & \textcolor{red}{\textbf{0.903}} &\textcolor{red}{\textbf{0.905}}\\ 
    \midrule
    \textit{Improvement to Fast-VQA} & - & \textit{+1.2\%}&\textit{+2.0\%} &\textit{+2.0\%}&\textit{+2.6\%} &\textit{+5.6\%}&\textit{+6.2\%}\\
    \bottomrule
  \end{tabular}
  \label{tab:transfer}
\end{table*}

\paragraph{\textbf{\textit{Newly-proposed LPVQ Dataset.}}} To validate the assessment of local perception, we present the first dataset encompassing local quality annotations, named as \textit{Local Perception Visual Quality (LPVQ)} dataset. 
Given the substantial cost of annotating videos, we build a dataset using images as static videos and conduct spatial-level annotations as a rational decision.
As videos are temporal extensions of images, our conclusions on visual saliency and local texture remain valid in images, allowing for the validation of local perception.
LPVQ comprises a total of 50 images meticulously collected from a typical short-form video platform, showcasing a wide range of scenes and quality factors to ensure representativeness, as depicted in Fig.~\ref{fig:lpvq}.

We evenly divide each image into non-overlapping $7\times 7$ grids. 
We assign a subjective quality rating ranging from 1 to 5 points (interval of 0.5) to each patch, involving 14 expert visual researchers for annotation.
After a glance at the entire image, all participants sequentially score each patch while other patches are occluded.
LPVQ comprises a total of \textit{34,300 annotations}, ensuring reliability.
Dataset details are elaborated in the supplementary materials.



\paragraph{\textbf{\textit{Implementation Details.}}} We use the Video Swin-T Tiny \cite{DBLP:conf/cvpr/LiuN0W00022} pretrained on Kinetics-400 \cite{DBLP:journals/corr/KayCSZHVVGBNSZ17} as the backbone before training on VQA tasks. The window size is $[8,7,7]$. 
Following the segment-based sampling strategy widely adopted \cite{DBLP:conf/eccv/WuCHLWSYL22,DBLP:journals/tip/TuWBAB21,DBLP:journals/corr/abs-2210-05357}, 32 discrete frames are extracted from 8 uniformly non-overlapping segments divided in each video included in the training and validation corpora. 
To maintain more complete semantic information, we resort to resizing each frame to a uniform resolution of $448 \times 448$. 
All other details are elaborated in the supplementary materials.


\subsection{Comparison with SOTA Results}
In Tab.~\ref{tab:sota}, we report the SRCC and PLCC with current SOTA methods on intra-dataset and generalization (\textit{i.e.}, cross-database evaluation) settings.
KVQ surpasses classical methods (\textit{e.g.}, TLVQM \cite{DBLP:journals/tip/Korhonen19}, VIDEVAL \cite{DBLP:journals/tip/TuWBAB21}) that rely on hand-crafted features in large margins. 
Compared with CNN-based methods (\textit{e.g.}, VSFA \cite{DBLP:conf/mm/LiJJ19}, PVQ \cite{DBLP:conf/cvpr/YingMGB21}, and Li \textit{et al.} \cite{DBLP:journals/tcsv/LiZTZW22}), KVQ showcases the advantage of attention mechanisms and maintains a significant lead in both intra-dataset and cross-dataset testing.
Compared to the current best models, Fast-VQA \cite{DBLP:conf/eccv/WuCHLWSYL22} and its variant Faster-VQA \cite{DBLP:journals/corr/abs-2210-05357}, which are also based on the Swin-T architecture, our model performs better through structural and constraint improvements. 
\textbf{We attained the best SRCC of 0.896 (+2.3\%) and PLCC of 0.897 (+2.3\%) in LSVQ\textsubscript{test}. In LSVQ\textsubscript{1080p}, our SRCC improved by 4.5\% to reach 0.814, while our PLCC increased by 3.9\% to reach 0.846.}
During the cross-database evaluation, KVQ improved by 4\% on koNViD-1k while maintaining a comparable performance on LIVE-VQC, thus exemplifying its generalization ability.


\paragraph{\textbf{\textit{Transfer learning on smaller VQA benchmarks.}}}
Following common practice \cite{DBLP:conf/eccv/WuCHLWSYL22}, we split all the datasets into 80\% training videos and 20\% testing videos randomly 10 times and report the average results.
In Tab.~\ref{tab:transfer}, KVQ continues to achieve superior results compared to existing SOTA algorithms on three benchmarks. \textbf{Our experimentation reveals a notable enhancement of approximately 2\% in performance on both LIVE-VQC and KoNViD-1k. Moreover, we achieve a remarkable improvement of 5\% to 6\% on YouTube-UGC}. 
KVQ utilizes the Swin-T based architecture and incorporates the FWA module to establish global correlations of both temporal and spatial domains. 
In comparison to other ViT-based methods, such as StarVQA \cite{xing2022starvqa} and VQT \cite{DBLP:conf/mm/YuanKZSW23} based on TimeSformer \cite{bertasius2021space}, KVQ demonstrates superior effectiveness.
Due to the effective modeling of saliency and local texture, KVQ demonstrates strong potential in transfer learning.


\subsection{Ablation Studies}

\paragraph{\textbf{\textit{Effectiveness of the FWA module.}} }
We compare the results of applying FWA with two other variants: applying only the IWA (\textit{i.e.}, the original Swin-T backbone) and applying only the CWA.
In Tab.~\ref{tab:FWA}, the results of applying only CWA are the poorest. 
The semantics of patches rely on neighboring regions for comprehension. Thus, solely using CWA leads to losing this crucial relationship.
Applying FWA yields superior results compared to solely applying IWA, particularly on high-resolution datasets LSVQ\textsubscript{1080p}.
High-resolution videos often encompass richer content. 
CWA enables the allocation of long-range attention, capturing more accurate visual saliency.

\begin{table}[t]
  \centering
  \footnotesize
  \caption{Ablation study on the proposed FWA module.}
  \label{tab:FWA}
  \resizebox{1.0\linewidth}{!}{
  \begin{tabular}{c|c|c|c|c}
    \toprule
    \multirow{2}{*}{Methods}  & LSVQ\textsubscript{test} & LSVQ\textsubscript{1080p} & KoNViD-1k & LIVE-VQC  \\
    & SRCC/PLCC & SRCC/PLCC & SRCC/PLCC & SRCC/PLCC  \\ 
    \midrule
    IWA & 0.894/0.894 & 0.807/0.839 & 0.886/0.888 & 0.816/0.839\\
    CWA & 0.881/0.882 & 0.793/0.825 & 0.871/0.873 & 0.801/0.833\\
    \midrule
    FWA & \textbf{0.896}/\textbf{0.897} & \textbf{0.814}/\textbf{0.846} & \textbf{0.890}/\textbf{0.892} & \textbf{0.820}/\textbf{0.843}\\
    \bottomrule
  \end{tabular}}
\end{table}
\begin{figure}[t]
  \centering
    \includegraphics[width=0.96\linewidth]{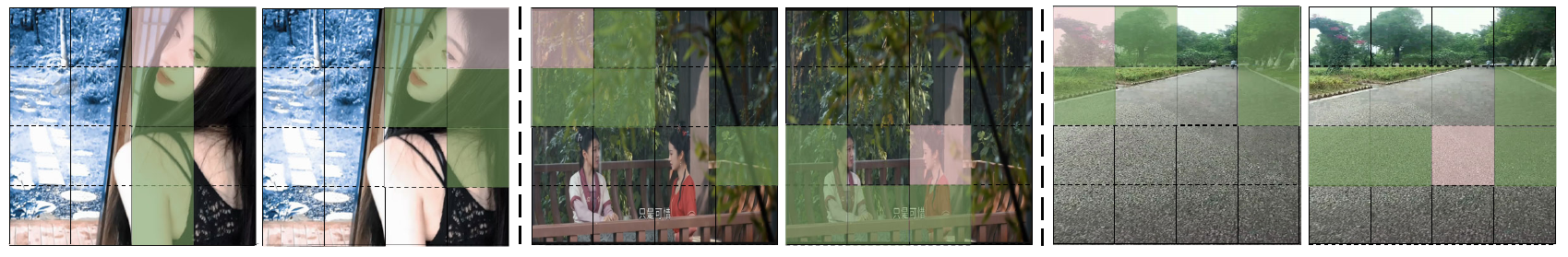}
  \caption{Visualization of the correlated windows, where \textbf{\textcolor{red}{red}} regions refer to the selected windows and \textbf{\textcolor[rgb]{0,0.5,0}{green}} regions refer to the correlated windows that are most related to them.}
  \label{fig:window}
\end{figure}
In Fig.~\ref{fig:window}, we visualize the selection results of the correlated windows in FWA, inputting images from the LPVQ dataset.
We can observe that FWA accurately models the semantics, as regions with the same semantics can be correctly associated.
Even in non-adjacent regions (\textit{e.g.}, trees on both sides, two people sitting apart), FWA can establish cross-window attention. 
This indicates that our FWA is capable of capturing long-range correlation between objects.


\paragraph{\textbf{\textit{Effectiveness of the Local Perception Constraint.}}}
We compare the results before and after adding the LPC. As shown in Tab.~\ref{tab:LPC}, the incorporation of the constraint has led to improvements in the results, particularly in cross-dataset evaluations. 
This signifies an enhancement in the generalization capabilities of the model.
The low-quality factors vary across different datasets, and applying the LPC enables a more discerning detection of these factors.

\begin{table}[t]
  \centering
  \footnotesize
  \caption{Ablation study on the proposed LPC.}
  \label{tab:LPC}
  \resizebox{1.0\linewidth}{!}{
  \begin{tabular}{c|c|c|c|c}
    \toprule
    \multirow{2}{*}{Methods}  & LSVQ\textsubscript{test} & LSVQ\textsubscript{1080p} & KoNViD-1k & LIVE-VQC  \\
    & SRCC/PLCC & SRCC/PLCC & SRCC/PLCC & SRCC/PLCC  \\ 
    \midrule
    w/o LPC & 0.894/0.895 & 0.811/0.842 & 0.887/0.888 & 0.804/0.835\\
    w LPC & \textbf{0.896}/\textbf{0.897} & \textbf{0.814}/\textbf{0.846} & \textbf{0.890}/\textbf{0.892} & \textbf{0.820}/\textbf{0.843}\\
    \bottomrule
  \end{tabular}}
\end{table}

\subsection{Local Perception and Saliency Analysis}
\paragraph{\textbf{\textit{Local Perception on LPVQ Dataset.}}}{
To accommodate the model's input requirements, each image is replicated as a 16-frame video and fed into the trained network for validation.
We compare KVQ with the existing SOTA method, Fast-VQA, which can also generate local prediction maps.
To validate the effectiveness of the LPC, we compare the results with and without the inclusion of LPC.
We compute the PLCC/SRCC for all annotations across the dataset as the inter-sample evaluation and calculate the average SRCC/PLCC for the intra-sample evaluation to validate the monotonicity of regional perception within an image.

\begin{table}[t]
  \centering
  \footnotesize
  \caption{Performance on the proposed LPVQ dataset.}
  \resizebox{0.76\linewidth}{!}{
  \begin{tabular}{c|c|c|c|c}
    \toprule
    \multirow{2}{*}{Method} & \multicolumn{2}{c|}{Inter-sample} & \multicolumn{2}{c}{Intra-sample}\\
    & SRCC & PLCC& SRCC & PLCC \\ 
    \midrule
    Fast-VQA  & 0.389 &0.407 & 0.280 & 0.289 \\
    KVQ (\textit{w/o LPC}) &0.405 &0.386&0.322&0.310\\
    KVQ (\textit{w LPC}) &\textbf{0.614} &\textbf{0.616}&\textbf{0.612}&\textbf{0.657}\\
    
    \bottomrule
  \end{tabular}}
  \label{tab:lpvq}
\end{table}
}
In Tab.~\ref{tab:lpvq}, our method outperforms Fast-VQA by a significant margin in both inter- and intra-sample evaluations.
This shows the capability of KVQ to assess the local quality accurately.
Our \textit{Assumption \ref{thm:local}} states that the local texture of each region is relatively independent. 
Therefore, we propose LPC to allow the perceptual predictions to focus more on the internal texture and distortion within each region. 
The noticeable performance improvement after incorporating the LPC stems from and validates this assumption.

\paragraph{\textbf{\textit{Saliency Assessment on SALICON Dataset.}}}{
The annotations of LPVQ are used to evaluate \textbf{local textures} without the ability to verify \textbf{saliency}.
We use the widely-used saliency evaluation dataset SALICON \cite{jiang2015salicon}
to verify the saliency prediction $\mathcal{S}$ with ground truth $\mathcal{S}_{gt}$. 
In addition to the \textbf{SRCC/PLCC}, we also employ the metrics as.
\begin{itemize}
    \item \textbf{sAUC/NSS.}  Regions that satisfy $\mathcal{S}_{GT}>0.9max(\mathcal{S}_{GT})$ are labeled as fixation points, and the classic \textbf{sAUC} and \textbf{NSS} metrics in saliency prediction field \cite{jia2020revisiting} are adopted.
    \item \textbf{KL.} 
    We compute the \textbf{KL divergence} to evaluate the similarity between the predictions and the ground truth.
\end{itemize}
Due to the particularity of saliency prediction tasks, as depicted in Fig.~\ref{fig:visualization}, the ground truth for salient regions is highly concentrated at fixation points.
In Tab.~\ref{tab:saliency}, as a completely untrained model for saliency prediction tasks, KVQ achieves remarkably high accuracy.

\begin{table}[t]
  \centering
  \footnotesize
  \caption{Saliency Assessment on the SALICON Dataset.}
    \resizebox{0.68\linewidth}{!}{
                  \begin{tabular}{c|ccccc}
                    \toprule
                     Methods  & SRCC & PLCC & sAUC & NSS & KL \\
                    \midrule
                    KVQ & 0.667 & 0.601 &0.850 & 1.384 & 0.607\\
                  \bottomrule
                  \end{tabular}}
  \label{tab:saliency}
\end{table}
}

\paragraph{\textbf{\textit{Visualization results.}}}{
\begin{figure}[t]
  \centering
    \includegraphics[width=0.95\linewidth]{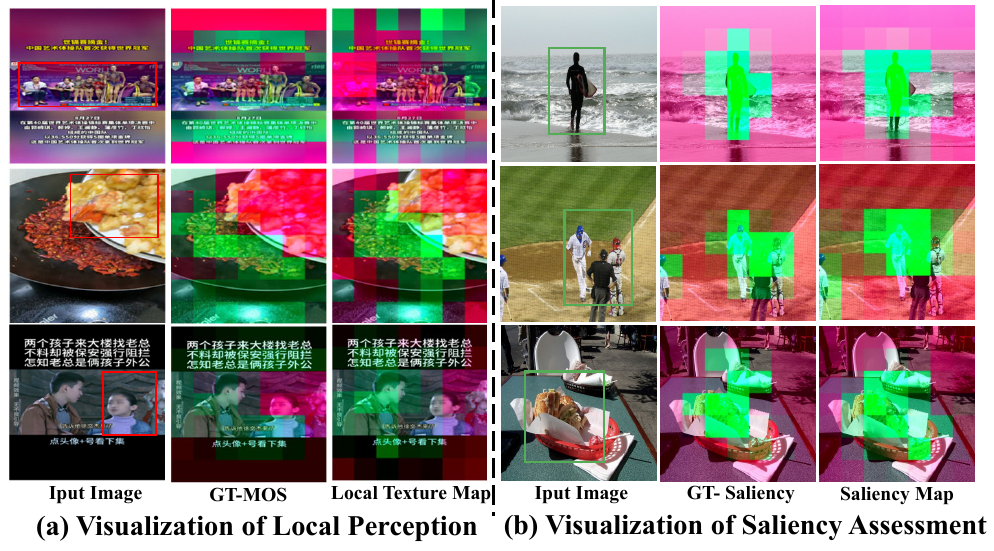}
  \caption{Visualization of the predicted maps, where \textbf{\textcolor{red}{red}} regions refer to low quality or low saliency and \textbf{\textcolor[rgb]{0,0.5,0}{green}} regions refer to high quality or high saliency (Please zoom in).}
  \label{fig:visualization}
\end{figure}
We visualize the predicted local texture maps and the ground truth MOS distribution on the LPVQ dataset in Fig.~\ref{fig:visualization}(a).
The predictions are generally similar to the ground truth and human perception. 
For instance, in the first row, the central portrait in the image, being relatively blurry, is accurately captured in the model's prediction.
In rows 2 and 3, the low-quality areas affected by motion blur receive lower texture scores, while the visually clearer regions receive higher texture scores.

In Fig.~\ref{fig:visualization}(b), we display the predictions of saliency maps and the ground truth.
Our KVQ is capable of highlighting foreground salient regions that contain more informative content, which is consistent with human observation.
}

\section{Conclusion}
To utilize the local perception of different regions, we analyzed the HVS principles and proposed the KVQ framework, which uses the FWA for attention allocation and the LPC for mitigating neighboring reliance. 
It achieved SOTA results across intra-dataset, cross-dataset, and transfer learning scenarios. 
To validate local perception, we established the first LPVQ dataset with region-wise annotations.
Experimental results on the LPVQ demonstrate our capacity.

\section*{Acknowledgments}

This paper is supported by BNRist projects (No. BNR20231880004 and No.BNR2024TD03003) and cash and in-kind contributions from the industry partner(s).

{
    \small
    \bibliographystyle{ieeenat_fullname}
    \bibliography{main}

\begin{thebibliography}{65}
\providecommand{\natexlab}[1]{#1}
\providecommand{\url}[1]{\texttt{#1}}
\expandafter\ifx\csname urlstyle\endcsname\relax
  \providecommand{\doi}[1]{doi: #1}\else
  \providecommand{\doi}{doi: \begingroup \urlstyle{rm}\Url}\fi

\bibitem[Bertasius et~al.(2021)Bertasius, Wang, and Torresani]{bertasius2021space}
Gedas Bertasius, Heng Wang, and Lorenzo Torresani.
\newblock Is space-time attention all you need for video understanding?
\newblock In \emph{ICML}, page~4, 2021.

\bibitem[Bosse et~al.(2018)Bosse, Maniry, M{\"{u}}ller, Wiegand, and Samek]{DBLP:journals/tip/BosseMMWS18}
Sebastian Bosse, Dominique Maniry, Klaus{-}Robert M{\"{u}}ller, Thomas Wiegand, and Wojciech Samek.
\newblock Deep neural networks for no-reference and full-reference image quality assessment.
\newblock \emph{{IEEE} Trans. Image Process.}, 27\penalty0 (1):\penalty0 206--219, 2018.

\bibitem[Bruce and Tsotsos(2005)]{bruce2005saliency}
Neil Bruce and John Tsotsos.
\newblock Saliency based on information maximization.
\newblock \emph{Advances in neural information processing systems}, 18, 2005.

\bibitem[Chadha and Andreopoulos(2021)]{DBLP:conf/cvpr/ChadhaA21}
Aaron Chadha and Yiannis Andreopoulos.
\newblock Deep perceptual preprocessing for video coding.
\newblock In \emph{{CVPR}}, pages 14852--14861. Computer Vision Foundation / {IEEE}, 2021.

\bibitem[Chen et~al.(2023)Chen, Mo, Hou, Wu, Liao, Sun, Yan, and Lin]{DBLP:journals/corr/topiq}
Chaofeng Chen, Jiadi Mo, Jingwen Hou, Haoning Wu, Liang Liao, Wenxiu Sun, Qiong Yan, and Weisi Lin.
\newblock {TOPIQ:} {A} top-down approach from semantics to distortions for image quality assessment.
\newblock \emph{CoRR}, abs/2308.03060, 2023.

\bibitem[Ciancio et~al.(2011)Ciancio, da~Costa, da~Silva, Said, Samadani, and Obrador]{DBLP:journals/tip/CiancioCSSSO11}
Alexandre~G. Ciancio, Andr{\'{e}} Luiz N.~Targino da Costa, Eduardo A.~B. da Silva, Amir Said, Ramin Samadani, and Pere Obrador.
\newblock No-reference blur assessment of digital pictures based on multifeature classifiers.
\newblock \emph{{IEEE} Trans. Image Process.}, 20\penalty0 (1):\penalty0 64--75, 2011.

\bibitem[Deng et~al.(2009)Deng, Dong, Socher, Li, Li, and Fei{-}Fei]{DBLP:conf/cvpr/DengDSLL009}
Jia Deng, Wei Dong, Richard Socher, Li{-}Jia Li, Kai Li, and Li Fei{-}Fei.
\newblock Imagenet: {A} large-scale hierarchical image database.
\newblock In \emph{{CVPR}}, pages 248--255. {IEEE} Computer Society, 2009.

\bibitem[Fang et~al.(2020)Fang, Zhu, Zeng, Ma, and Wang]{DBLP:conf/cvpr/FangZZMW20}
Yuming Fang, Hanwei Zhu, Yan Zeng, Kede Ma, and Zhou Wang.
\newblock Perceptual quality assessment of smartphone photography.
\newblock In \emph{{CVPR}}, pages 3674--3683. Computer Vision Foundation / {IEEE}, 2020.

\bibitem[Ghadiyaram and Bovik(2016)]{DBLP:journals/tip/GhadiyaramB16}
Deepti Ghadiyaram and Alan~C. Bovik.
\newblock Massive online crowdsourced study of subjective and objective picture quality.
\newblock \emph{{IEEE} Trans. Image Process.}, 25\penalty0 (1):\penalty0 372--387, 2016.

\bibitem[Gu et~al.(2016)Gu, Wang, Yang, Lin, Zhai, Yang, and Zhang]{DBLP:journals/tmm/GuWYLZY016}
Ke Gu, Shiqi Wang, Huan Yang, Weisi Lin, Guangtao Zhai, Xiaokang Yang, and Wenjun Zhang.
\newblock Saliency-guided quality assessment of screen content images.
\newblock \emph{{IEEE} Trans. Multim.}, 18\penalty0 (6):\penalty0 1098--1110, 2016.

\bibitem[Guan et~al.(2017)Guan, Yi, Zeng, Cham, and Wang]{DBLP:journals/tmm/GuanYZCW17}
Jingwei Guan, Shuai Yi, Xingyu Zeng, Wai{-}Kuen Cham, and Xiaogang Wang.
\newblock Visual importance and distortion guided deep image quality assessment framework.
\newblock \emph{{IEEE} Trans. Multim.}, 19\penalty0 (11):\penalty0 2505--2520, 2017.

\bibitem[Haralick et~al.(1973)Haralick, Shanmugam, and Dinstein]{haralick1973textural}
Robert~M Haralick, Karthikeyan Shanmugam, and Its'~Hak Dinstein.
\newblock Textural features for image classification.
\newblock \emph{IEEE Transactions on systems, man, and cybernetics}, \penalty0 (6):\penalty0 610--621, 1973.

\bibitem[Hosu et~al.(2017)Hosu, Hahn, Jenadeleh, Lin, Men, Szir{\'{a}}nyi, Li, and Saupe]{DBLP:conf/qomex/HosuHJLMSLS17}
Vlad Hosu, Franz Hahn, Mohsen Jenadeleh, Hanhe Lin, Hui Men, Tam{\'{a}}s Szir{\'{a}}nyi, Shujun Li, and Dietmar Saupe.
\newblock The konstanz natural video database (konvid-1k).
\newblock In \emph{QoMEX}, pages 1--6. {IEEE}, 2017.

\bibitem[Hosu et~al.(2019)Hosu, Goldl{\"{u}}cke, and Saupe]{DBLP:conf/cvpr/HosuGS19}
Vlad Hosu, Bastian Goldl{\"{u}}cke, and Dietmar Saupe.
\newblock Effective aesthetics prediction with multi-level spatially pooled features.
\newblock In \emph{{IEEE} Conference on Computer Vision and Pattern Recognition, {CVPR} 2019, Long Beach, CA, USA, June 16-20, 2019}, pages 9375--9383. Computer Vision Foundation / {IEEE}, 2019.

\bibitem[Hosu et~al.(2020)Hosu, Lin, Szir{\'{a}}nyi, and Saupe]{DBLP:journals/tip/HosuLSS20}
Vlad Hosu, Hanhe Lin, Tam{\'{a}}s Szir{\'{a}}nyi, and Dietmar Saupe.
\newblock Koniq-10k: An ecologically valid database for deep learning of blind image quality assessment.
\newblock \emph{{IEEE} Trans. Image Process.}, 29:\penalty0 4041--4056, 2020.

\bibitem[Jia and Bruce(2020)]{jia2020revisiting}
Sen Jia and Neil~DB Bruce.
\newblock Revisiting saliency metrics: Farthest-neighbor area under curve.
\newblock In \emph{Proceedings of the IEEE/CVF Conference on Computer Vision and Pattern Recognition}, pages 2667--2676, 2020.

\bibitem[Jiang et~al.(2015)Jiang, Huang, Duan, and Zhao]{jiang2015salicon}
Ming Jiang, Shengsheng Huang, Juanyong Duan, and Qi Zhao.
\newblock Salicon: Saliency in context.
\newblock In \emph{The IEEE Conference on Computer Vision and Pattern Recognition (CVPR)}, 2015.

\bibitem[Kay et~al.(2017)Kay, Carreira, Simonyan, Zhang, Hillier, Vijayanarasimhan, Viola, Green, Back, Natsev, Suleyman, and Zisserman]{DBLP:journals/corr/KayCSZHVVGBNSZ17}
Will Kay, Jo{\~{a}}o Carreira, Karen Simonyan, Brian Zhang, Chloe Hillier, Sudheendra Vijayanarasimhan, Fabio Viola, Tim Green, Trevor Back, Paul Natsev, Mustafa Suleyman, and Andrew Zisserman.
\newblock The kinetics human action video dataset.
\newblock \emph{CoRR}, abs/1705.06950, 2017.

\bibitem[Ke et~al.(2021)Ke, Wang, Wang, Milanfar, and Yang]{DBLP:conf/iccv/KeWWMY21}
Junjie Ke, Qifei Wang, Yilin Wang, Peyman Milanfar, and Feng Yang.
\newblock {MUSIQ:} multi-scale image quality transformer.
\newblock In \emph{2021 {IEEE/CVF} International Conference on Computer Vision, {ICCV} 2021, Montreal, QC, Canada, October 10-17, 2021}, pages 5128--5137. {IEEE}, 2021.

\bibitem[Korhonen(2019)]{DBLP:journals/tip/Korhonen19}
Jari Korhonen.
\newblock Two-level approach for no-reference consumer video quality assessment.
\newblock \emph{{IEEE} Trans. Image Process.}, 28\penalty0 (12):\penalty0 5923--5938, 2019.

\bibitem[Li et~al.(2022)Li, Zhang, Tian, Zhai, and Wang]{DBLP:journals/tcsv/LiZTZW22}
Bowen Li, Weixia Zhang, Meng Tian, Guangtao Zhai, and Xianpei Wang.
\newblock Blindly assess quality of in-the-wild videos via quality-aware pre-training and motion perception.
\newblock \emph{{IEEE} Trans. Circuits Syst. Video Technol.}, 32\penalty0 (9):\penalty0 5944--5958, 2022.

\bibitem[Li et~al.(2019)Li, Jiang, and Jiang]{DBLP:conf/mm/LiJJ19}
Dingquan Li, Tingting Jiang, and Ming Jiang.
\newblock Quality assessment of in-the-wild videos.
\newblock In \emph{{ACM} Multimedia}, pages 2351--2359. {ACM}, 2019.

\bibitem[Li et~al.(2020)Li, Jiang, and Jiang]{DBLP:conf/mm/LiJJ20}
Dingquan Li, Tingting Jiang, and Ming Jiang.
\newblock Norm-in-norm loss with faster convergence and better performance for image quality assessment.
\newblock In \emph{{ACM} Multimedia}, pages 789--797. {ACM}, 2020.

\bibitem[Li et~al.(2021{\natexlab{a}})Li, Jiang, and Jiang]{DBLP:journals/ijcv/LiJJ21}
Dingquan Li, Tingting Jiang, and Ming Jiang.
\newblock Unified quality assessment of in-the-wild videos with mixed datasets training.
\newblock \emph{Int. J. Comput. Vis.}, 129\penalty0 (4):\penalty0 1238--1257, 2021{\natexlab{a}}.

\bibitem[Li et~al.(2021{\natexlab{b}})Li, Zhang, and Cosman]{DBLP:journals/tcsv/LiZC21}
Fan Li, Yangfan Zhang, and Pamela~C. Cosman.
\newblock Mmmnet: An end-to-end multi-task deep convolution neural network with multi-scale and multi-hierarchy fusion for blind image quality assessment.
\newblock \emph{{IEEE} Trans. Circuits Syst. Video Technol.}, 31\penalty0 (12):\penalty0 4798--4811, 2021{\natexlab{b}}.

\bibitem[Liu et~al.(2023)Liu, Wu, Yuan, Sun, Tang, Zheng, Wen, and Li]{liu2023ada}
Hongbo Liu, Mingda Wu, Kun Yuan, Ming Sun, Yansong Tang, Chuanchuan Zheng, Xing Wen, and Xiu Li.
\newblock Ada-dqa: Adaptive diverse quality-aware feature acquisition for video quality assessment.
\newblock In \emph{Proceedings of the 31st ACM International Conference on Multimedia}, pages 6695--6704, 2023.

\bibitem[Liu et~al.(2021)Liu, Lin, Cao, Hu, Wei, Zhang, Lin, and Guo]{DBLP:conf/iccv/LiuL00W0LG21}
Ze Liu, Yutong Lin, Yue Cao, Han Hu, Yixuan Wei, Zheng Zhang, Stephen Lin, and Baining Guo.
\newblock Swin transformer: Hierarchical vision transformer using shifted windows.
\newblock In \emph{{ICCV}}, pages 9992--10002. {IEEE}, 2021.

\bibitem[Liu et~al.(2022)Liu, Ning, Cao, Wei, Zhang, Lin, and Hu]{DBLP:conf/cvpr/LiuN0W00022}
Ze Liu, Jia Ning, Yue Cao, Yixuan Wei, Zheng Zhang, Stephen Lin, and Han Hu.
\newblock Video swin transformer.
\newblock In \emph{{CVPR}}, pages 3192--3201. {IEEE}, 2022.

\bibitem[Lu et~al.(2024)Lu, Li, Pei, Yuan, Xie, Qu, Sun, Zhou, and Chen]{lu2024kvq}
Yiting Lu, Xin Li, Yajing Pei, Kun Yuan, Qizhi Xie, Yunpeng Qu, Ming Sun, Chao Zhou, and Zhibo Chen.
\newblock Kvq: Kwai video quality assessment for short-form videos.
\newblock In \emph{Proceedings of the IEEE/CVF Conference on Computer Vision and Pattern Recognition}, pages 25963--25973, 2024.

\bibitem[Ma et~al.(2018)Ma, Liu, Zhang, Duanmu, Wang, and Zuo]{DBLP:journals/tip/MaLZDWZ18}
Kede Ma, Wentao Liu, Kai Zhang, Zhengfang Duanmu, Zhou Wang, and Wangmeng Zuo.
\newblock End-to-end blind image quality assessment using deep neural networks.
\newblock \emph{{IEEE} Trans. Image Process.}, 27\penalty0 (3):\penalty0 1202--1213, 2018.

\bibitem[Mittal et~al.(2012)Mittal, Moorthy, and Bovik]{DBLP:journals/tip/MittalMB12}
Anish Mittal, Anush~Krishna Moorthy, and Alan~Conrad Bovik.
\newblock No-reference image quality assessment in the spatial domain.
\newblock \emph{{IEEE} Trans. Image Process.}, 21\penalty0 (12):\penalty0 4695--4708, 2012.

\bibitem[Sinno and Bovik(2019)]{DBLP:journals/tip/SinnoB19}
Zeina Sinno and Alan~Conrad Bovik.
\newblock Large-scale study of perceptual video quality.
\newblock \emph{{IEEE} Trans. Image Process.}, 28\penalty0 (2):\penalty0 612--627, 2019.

\bibitem[Sun et~al.(2022)Sun, Min, Lu, and Zhai]{DBLP:conf/mm/SunMLZ22}
Wei Sun, Xiongkuo Min, Wei Lu, and Guangtao Zhai.
\newblock A deep learning based no-reference quality assessment model for {UGC} videos.
\newblock In \emph{{MM} '22: The 30th {ACM} International Conference on Multimedia, Lisboa, Portugal, October 10 - 14, 2022}, pages 856--865. {ACM}, 2022.

\bibitem[Thomee et~al.(2016)Thomee, Shamma, Friedland, Elizalde, Ni, Poland, Borth, and Li]{DBLP:journals/cacm/ThomeeSFENPBL16}
Bart Thomee, David~A. Shamma, Gerald Friedland, Benjamin Elizalde, Karl Ni, Douglas Poland, Damian Borth, and Li{-}Jia Li.
\newblock {YFCC100M:} the new data in multimedia research.
\newblock \emph{Commun. {ACM}}, 59\penalty0 (2):\penalty0 64--73, 2016.

\bibitem[Treue(2003)]{treue2003visual}
Stefan Treue.
\newblock Visual attention: the where, what, how and why of saliency.
\newblock \emph{Current opinion in neurobiology}, 13\penalty0 (4):\penalty0 428--432, 2003.

\bibitem[Tu et~al.(2021{\natexlab{a}})Tu, Wang, Birkbeck, Adsumilli, and Bovik]{DBLP:journals/tip/TuWBAB21}
Zhengzhong Tu, Yilin Wang, Neil Birkbeck, Balu Adsumilli, and Alan~C. Bovik.
\newblock {UGC-VQA:} benchmarking blind video quality assessment for user generated content.
\newblock \emph{{IEEE} Trans. Image Process.}, 30:\penalty0 4449--4464, 2021{\natexlab{a}}.

\bibitem[Tu et~al.(2021{\natexlab{b}})Tu, Yu, Wang, Birkbeck, Adsumilli, and Bovik]{DBLP:journals/corr/rapique}
Zhengzhong Tu, Xiangxu Yu, Yilin Wang, Neil Birkbeck, Balu Adsumilli, and Alan~C. Bovik.
\newblock {RAPIQUE:} rapid and accurate video quality prediction of user generated content.
\newblock \emph{CoRR}, abs/2101.10955, 2021{\natexlab{b}}.

\bibitem[Vaswani et~al.(2017)Vaswani, Shazeer, Parmar, Uszkoreit, Jones, Gomez, Kaiser, and Polosukhin]{DBLP:conf/nips/VaswaniSPUJGKP17}
Ashish Vaswani, Noam Shazeer, Niki Parmar, Jakob Uszkoreit, Llion Jones, Aidan~N. Gomez, Lukasz Kaiser, and Illia Polosukhin.
\newblock Attention is all you need.
\newblock In \emph{{NIPS}}, pages 5998--6008, 2017.

\bibitem[Wang et~al.(2022)Wang, Yao, Chen, Lin, Cai, He, and Liu]{DBLP:conf/iclr/0001YCL00L22}
Wenxiao Wang, Lu Yao, Long Chen, Binbin Lin, Deng Cai, Xiaofei He, and Wei Liu.
\newblock Crossformer: {A} versatile vision transformer hinging on cross-scale attention.
\newblock In \emph{{ICLR}}. OpenReview.net, 2022.

\bibitem[Wang et~al.(2019)Wang, Inguva, and Adsumilli]{DBLP:conf/mmsp/WangIA19}
Yilin Wang, Sasi Inguva, and Balu Adsumilli.
\newblock Youtube {UGC} dataset for video compression research.
\newblock In \emph{{MMSP}}, pages 1--5. {IEEE}, 2019.

\bibitem[Wang et~al.(2021)Wang, Ke, Talebi, Yim, Birkbeck, Adsumilli, Milanfar, and Yang]{DBLP:conf/cvpr/WangKTYBAMY21}
Yilin Wang, Junjie Ke, Hossein Talebi, Joong~Gon Yim, Neil Birkbeck, Balu Adsumilli, Peyman Milanfar, and Feng Yang.
\newblock Rich features for perceptual quality assessment of {UGC} videos.
\newblock In \emph{{CVPR}}, pages 13435--13444. Computer Vision Foundation / {IEEE}, 2021.

\bibitem[Wang and Li(2010)]{wang2010information}
Zhou Wang and Qiang Li.
\newblock Information content weighting for perceptual image quality assessment.
\newblock \emph{IEEE Transactions on image processing}, 20\penalty0 (5):\penalty0 1185--1198, 2010.

\bibitem[Wang et~al.(2004)Wang, Bovik, Sheikh, and Simoncelli]{wang2004image}
Zhou Wang, Alan~C Bovik, Hamid~R Sheikh, and Eero~P Simoncelli.
\newblock Image quality assessment: from error visibility to structural similarity.
\newblock \emph{IEEE transactions on image processing}, 13\penalty0 (4):\penalty0 600--612, 2004.

\bibitem[Winkler(2012)]{DBLP:journals/jstsp/Winkler12}
Stefan Winkler.
\newblock Analysis of public image and video databases for quality assessment.
\newblock \emph{{IEEE} J. Sel. Top. Signal Process.}, 6\penalty0 (6):\penalty0 616--625, 2012.

\bibitem[Wu et~al.(2022{\natexlab{a}})Wu, Chen, Hou, Liao, Wang, Sun, Yan, and Lin]{DBLP:conf/eccv/WuCHLWSYL22}
Haoning Wu, Chaofeng Chen, Jingwen Hou, Liang Liao, Annan Wang, Wenxiu Sun, Qiong Yan, and Weisi Lin.
\newblock {FAST-VQA:} efficient end-to-end video quality assessment with fragment sampling.
\newblock In \emph{{ECCV} {(6)}}, pages 538--554. Springer, 2022{\natexlab{a}}.

\bibitem[Wu et~al.(2022{\natexlab{b}})Wu, Chen, Liao, Hou, Sun, Yan, Gu, and Lin]{DBLP:journals/corr/abs-2210-05357}
Haoning Wu, Chaofeng Chen, Liang Liao, Jingwen Hou, Wenxiu Sun, Qiong Yan, Jinwei Gu, and Weisi Lin.
\newblock Neighbourhood representative sampling for efficient end-to-end video quality assessment.
\newblock \emph{CoRR}, abs/2210.05357, 2022{\natexlab{b}}.

\bibitem[Wu et~al.(2023)Wu, Chen, Liao, Hou, Sun, Yan, and Lin]{DBLP:journals/tcsv/WuCLHSYL23}
Haoning Wu, Chaofeng Chen, Liang Liao, Jingwen Hou, Wenxiu Sun, Qiong Yan, and Weisi Lin.
\newblock Discovqa: Temporal distortion-content transformers for video quality assessment.
\newblock \emph{{IEEE} Trans. Circuits Syst. Video Technol.}, 33\penalty0 (9):\penalty0 4840--4854, 2023.

\bibitem[Wurtz et~al.(2000)Wurtz, Kandel, et~al.]{wurtz2000central}
Robert~H Wurtz, Eric~R Kandel, et~al.
\newblock Central visual pathways.
\newblock \emph{Principles of neural science}, 4:\penalty0 523--545, 2000.

\bibitem[Xia et~al.(2022)Xia, Pan, Song, Li, and Huang]{DBLP:conf/cvpr/XiaPSLH22}
Zhuofan Xia, Xuran Pan, Shiji Song, Li~Erran Li, and Gao Huang.
\newblock Vision transformer with deformable attention.
\newblock In \emph{{CVPR}}, pages 4784--4793. {IEEE}, 2022.

\bibitem[Xie et~al.(2024)Xie, Yuan, Qu, Wu, Sun, Zhou, and Zhu]{xie2024qpt}
Qizhi Xie, Kun Yuan, Yunpeng Qu, Mingda Wu, Ming Sun, Chao Zhou, and Jihong Zhu.
\newblock Qpt-v2: Masked image modeling advances visual scoring.
\newblock In \emph{Proceedings of the 32nd ACM International Conference on Multimedia}, pages 2709--2718, 2024.

\bibitem[Xing et~al.(2022)Xing, Wang, Wang, Li, and Zhu]{xing2022starvqa}
Fengchuang Xing, Yuan-Gen Wang, Hanpin Wang, Leida Li, and Guopu Zhu.
\newblock Starvqa: Space-time attention for video quality assessment.
\newblock In \emph{2022 IEEE International Conference on Image Processing (ICIP)}, pages 2326--2330. IEEE, 2022.

\bibitem[Xing et~al.(2023)Xing, Wang, Tang, Zhu, and Kwong]{xing2023starvqa+}
Fengchuang Xing, Yuan-Gen Wang, Weixuan Tang, Guopu Zhu, and Sam Kwong.
\newblock Starvqa+: co-training space-time attention for video quality assessment.
\newblock \emph{arXiv preprint arXiv:2306.12298}, 2023.

\bibitem[Xing et~al.(2024)Xing, Li, Wang, Zhu, and Cao]{xing2024clipvqa}
Fengchuang Xing, Mingjie Li, Yuan-Gen Wang, Guopu Zhu, and Xiaochun Cao.
\newblock Clipvqa: Video quality assessment via clip.
\newblock \emph{IEEE Transactions on Broadcasting}, 2024.

\bibitem[Yang et~al.(2019)Yang, Jiang, Lin, and Wang]{DBLP:conf/mm/YangJLW19}
Sheng Yang, Qiuping Jiang, Weisi Lin, and Yongtao Wang.
\newblock Sgdnet: An end-to-end saliency-guided deep neural network for no-reference image quality assessment.
\newblock In \emph{{ACM} Multimedia}, pages 1383--1391. {ACM}, 2019.

\bibitem[Ying et~al.(2020)Ying, Niu, Gupta, Mahajan, Ghadiyaram, and Bovik]{DBLP:conf/cvpr/YingNGMGB20}
Zhenqiang Ying, Haoran Niu, Praful Gupta, Dhruv Mahajan, Deepti Ghadiyaram, and Alan~C. Bovik.
\newblock From patches to pictures (paq-2-piq): Mapping the perceptual space of picture quality.
\newblock In \emph{{CVPR}}, pages 3572--3582. Computer Vision Foundation / {IEEE}, 2020.

\bibitem[Ying et~al.(2021)Ying, Mandal, Ghadiyaram, and Bovik]{DBLP:conf/cvpr/YingMGB21}
Zhenqiang Ying, Maniratnam Mandal, Deepti Ghadiyaram, and Alan~C. Bovik.
\newblock Patch-vq: 'patching up' the video quality problem.
\newblock In \emph{{CVPR}}, pages 14019--14029. Computer Vision Foundation / {IEEE}, 2021.

\bibitem[Yuan et~al.(2023)Yuan, Kong, Zheng, Sun, and Wen]{DBLP:conf/mm/YuanKZSW23}
Kun Yuan, Zishang Kong, Chuanchuan Zheng, Ming Sun, and Xing Wen.
\newblock Capturing co-existing distortions in user-generated content for no-reference video quality assessment.
\newblock In \emph{{ACM} Multimedia}, pages 1098--1107. {ACM}, 2023.

\bibitem[Yuan et~al.(2024)Yuan, Liu, Li, Sun, Sun, Gong, Hao, Zhou, and Tang]{yuan2024ptm}
Kun Yuan, Hongbo Liu, Mading Li, Muyi Sun, Ming Sun, Jiachao Gong, Jinhua Hao, Chao Zhou, and Yansong Tang.
\newblock Ptm-vqa: efficient video quality assessment leveraging diverse pretrained models from the wild.
\newblock In \emph{Proceedings of the IEEE/CVF Conference on Computer Vision and Pattern Recognition}, pages 2835--2845, 2024.

\bibitem[Zhang et~al.(2018)Zhang, Isola, Efros, Shechtman, and Wang]{DBLP:conf/cvpr/ZhangIESW18}
Richard Zhang, Phillip Isola, Alexei~A. Efros, Eli Shechtman, and Oliver Wang.
\newblock The unreasonable effectiveness of deep features as a perceptual metric.
\newblock In \emph{{CVPR}}, pages 586--595. Computer Vision Foundation / {IEEE} Computer Society, 2018.

\bibitem[Zhang and Liu(2017)]{DBLP:journals/tip/ZhangL17}
Wei Zhang and Hantao Liu.
\newblock Study of saliency in objective video quality assessment.
\newblock \emph{{IEEE} Trans. Image Process.}, 26\penalty0 (3):\penalty0 1275--1288, 2017.

\bibitem[Zhang et~al.(2016)Zhang, Borji, Wang, Callet, and Liu]{DBLP:journals/tnn/ZhangBWCL16}
Wei Zhang, Ali Borji, Zhou Wang, Patrick~Le Callet, and Hantao Liu.
\newblock The application of visual saliency models in objective image quality assessment: {A} statistical evaluation.
\newblock \emph{{IEEE} Trans. Neural Networks Learn. Syst.}, 27\penalty0 (6):\penalty0 1266--1278, 2016.

\bibitem[Zhao et~al.(2023{\natexlab{a}})Zhao, Yuan, Sun, Li, and Wen]{zhao2023quality}
Kai Zhao, Kun Yuan, Ming Sun, Mading Li, and Xing Wen.
\newblock Quality-aware pre-trained models for blind image quality assessment.
\newblock In \emph{Proceedings of the IEEE/CVF conference on computer vision and pattern recognition}, pages 22302--22313, 2023{\natexlab{a}}.

\bibitem[Zhao et~al.(2023{\natexlab{b}})Zhao, Yuan, Sun, and Wen]{DBLP:conf/cvpr/ZhaoYSW23}
Kai Zhao, Kun Yuan, Ming Sun, and Xing Wen.
\newblock Zoom-vqa: Patches, frames and clips integration for video quality assessment.
\newblock In \emph{{CVPR} Workshops}, pages 1302--1310. {IEEE}, 2023{\natexlab{b}}.

\bibitem[Zhu et~al.(2023)Zhu, Wang, Ke, Zhang, and Lau]{DBLP:conf/cvpr/ZhuWKZL23}
Lei Zhu, Xinjiang Wang, Zhanghan Ke, Wayne Zhang, and Rynson W.~H. Lau.
\newblock Biformer: Vision transformer with bi-level routing attention.
\newblock In \emph{{CVPR}}, pages 10323--10333. {IEEE}, 2023.

\bibitem[Zhu et~al.(2021)Zhu, Hou, Chen, Xie, Lu, and Che]{DBLP:conf/iccvw/ZhuHCXLC21}
Mengmeng Zhu, Guanqun Hou, Xinjia Chen, Jiaxing Xie, Haixian Lu, and Jun Che.
\newblock Saliency-guided transformer network combined with local embedding for no-reference image quality assessment.
\newblock In \emph{{ICCVW}}, pages 1953--1962. {IEEE}, 2021.

\end{thebibliography}
}


\end{document}